\documentclass[a4paper]{article}
\usepackage{algorithmic}
\usepackage{algorithm}
\usepackage{multirow}
\usepackage{INTERSPEECH2020}
\title{Introducing a Central African Primate Vocalisation Dataset for Automated Species Classification}

\name{Joeri A. Zwerts$^1$, Jelle Treep$^{2}$, Casper S. Kaandorp$^{2}$, Floor Meewis$^{1}$, Amparo C. Koot$^{1}$, Heysem Kaya$^3$}
\address{
$^1$ Department of Biology, Utrecht University, Utrecht, The Netherlands \\
$^2$ Information and Technology Services, Utrecht University, Utrecht, The Netherlands\\
$^3$ Department of Information and Computing Sciences, Utrecht University, Utrecht, The Netherlands
}

\email{
j.a.zwerts@uu.nl,
h.kaya@uu.nl
}

\begin{document}

\maketitle
\begin{abstract}
Automated classification of animal vocalisations is a potentially powerful wildlife monitoring tool. Training robust classifiers requires sizable annotated datasets, which are not easily recorded in the wild. To circumvent this problem, we recorded four primate species under semi-natural conditions in a wildlife sanctuary in Cameroon with the objective to train a classifier capable of detecting species in the wild. Here, we introduce the collected dataset, describe our approach and initial results of classifier development. To increase the efficiency of the annotation process, we condensed the recordings with an energy/change based automatic vocalisation detection. Segmenting the annotated chunks into training, validation and test sets, initial results reveal up to 82\% unweighted average recall (UAR) test set performance in four-class primate species classification. 

\end{abstract}

\section{Introduction}
\label{sec:intro}
Wildlife is declining at unprecedented rates, and monitoring trends in biodiversity is key to engage in effective conservation actions~\cite{almond2020living}. Using acoustic recordings to identify and count species is a promising non-invasive and cost-effective monitoring tool~\cite{sugai2019terrestrial}. This can be particularly useful in environments with limited visibility such as tropical forests, or for arboreal, shy or nocturnal species that are more easily heard than seen. Acoustic monitoring, especially in conjunction with other monitoring methods, 
has the potential to profoundly change ecological research by opening up new ways of studying community composition, species interactions and behavioral processes~\cite{buxton2018pairing}. For efficient analysis of audio recordings however, automated detection is pivotal. In addition to relieving a data processing bottleneck, machine learning methods allow for consistency in terms of quality, increasing the comparability and reproducibility of the output.

Training robust classifiers requires sizable amounts of annotated data, which can require substantial efforts to compile from natural forest recordings. To circumvent this problem, we recorded several primate species in a sanctuary in Cameroon, including chimpanzees (\textit{Pan troglodytes}, n=20), mandrills (\textit{Mandrillus sphinx}, n=17), red-capped mangabeys (\textit{Cercocebus torquatus}, n=6) and a mixed group of guenon species (\textit{Cercopithecus spp.}, n=20). The primates in the sanctuary live in semi-natural conditions with background noise that is somewhat, although not wholly, comparable to natural background noise. The ultimate  objective of these efforts is to train a classifier capable of detecting species in the wild. This may also provide insights into whether this approach, of using sanctuary recordings, can be used to train classifiers for other species as well, to aid in the development of cost-effective monitoring to meet modern conservation challenges.

In this paper, we present the dataset, the semi-automatic annotation process that we used to speed up the manual annotation process, and a benchmark species classification system. 


\subsection{Related Work}
\label{sec:bg}
Multiple studies have applied automatic acoustic monitoring for a variety of taxa including cetaceans~\cite{bittle2013review}, birds~\cite{priyadarshani2018automated}, bats~\cite{russo2016use}, insects~\cite{ganchev2007automatic}, amphibians~\cite{brauer2016comparison}, and forest elephants~\cite{wrege2017acoustic}. However, they have so far only been sporadically been used for primates~\cite{mielke2013method,heinicke2015assessing,fedurek2016sequential,turesson2016machine,clink2019application,enari2019evaluation}. A brief summary of recent works on classification of primate vocalisations is given in Table~\ref{tab:relatedwork}. We observe that Mel-Frequency Cepstral Coefficients (MFCC) are commonly used in classifying primate vocalisations, in most cases without other acoustic descriptors. In our study, we also use MFCCs (together with temporal delta coefficients) and combine them with RASTA-style Perceptual Linear Prediction Cepstral Coefficients. There are also off-the-shelf applications like Kaleidoscope Pro (Wildlife Acoustics, MA, USA) based on Hidden Markov Models that were used in recent works for call type classification of Japanese macaques (\textit{Macaca fuscata})~\cite{enari2019evaluation}.

\begin{table*}[ht]
  \centering
  \caption{Summary of recent works on automatic primate vocalisation classification. }
    \begin{tabular}{p{8em}p{8em}p{14em}p{8em}p{9em}}
     \toprule
    \textbf{Work} & \textbf{Task(s)} & \textbf{Species} & \textbf{Features} & \textbf{Classifiers} \\ \midrule
    Mielke et al.~\cite{mielke2013method} & Three recognition tasks (individual, call type and species) & Blue monkey (\textit{Cercopithecus mitis stuhlmanni}), Olive baboon (\textit{Papio\newline{}anubis)}, Redtail monkey (\textit{Cercopithecus ascanius schmidti}), Guereza colobus (\textit{Colobus guereza occidentalis}) & MFCC [1-32]  and Deltas & MLP \\ \hline
    Heinicke et al.~\cite{heinicke2015assessing} & 5-class primate classification & Chimpanzee (\textit{Pan troglodytes}), Diana monkey (\textit{Cercopithecus diana}), King colobus (\textit{Colobus polykomos}) and Western red colobus (\textit{Procolobus badius}) & MFCCs, loudness, spectral crest factor, spectral flatness measure, and ZCR & SVM and GMM \\ \hline
    Fedurek et al.~\cite{fedurek2016sequential} & Age, Context, Identity, Social Status & Chimpanzee (\textit{Pan troglodytes}) & MFCCs & SVM \\ \hline
     Turesson et al.~\cite{turesson2016machine} & 8-class classification of Marmoset vocalisations & Common marmoset (\textit{Callithrix jacchus}) & LPC with LPF orders of 10, 15, 20 and 25 & AdaBoost, Bayesian Classifier,  k-NN, Logistic regression, MLP, SVM, Optimum Path Forest  \\ \hline
    Clink et al.~\cite{clink2019application} & Distinguishing individuals & Bornean gibbon (\textit{Hylobatidae muelleri}) & MFCC [1-12] & SVM \\ \bottomrule
    \end{tabular}%
  \label{tab:relatedwork}%
\end{table*}%

\section{Central African Primate Dataset}
\label{sec:appro}

\subsection{Acoustic Data Collection}
\label{subseq:datacoll}
The acoustic data is collected in the Mefou Primate Sanctuary (Ape Action Africa) in Cameroon in December 2019 and January 2020. The sanctuary, which houses the primates in a semi-natural forest setting, cares for rescued primates and engages in conservation and education initiatives. Recordings were made using Audiomoth (v1.1.0) recorders~\cite{hill2019audiomoth}. Devices recorded 1-min segments continuously at 48 kHz and 30.6 dB gain, storing the data in one minute WAVE-files, with interruptions from two to five seconds between recordings for the recorder to save the files. For all species, the recorders were installed either directly on the fence of their respective enclosures, or maximally up to 3 meters away from it. Per species, the enclosures differed in size and were approximately 40 $\times$ 40 meters in size for the guenons and red-capped mangabeys, 50 $\times$ 50 meters for the mandrills and 70 $\times$ 70 meters for the chimpanzees. Distance between the recorder and the animals naturally varied depending on the location of the animal within the enclosure. The smallest distance between two enclosures having different species was 30 meters. Due to the limited distance between some of the enclosures and the loudness of the vocalisations, some level of interference (i.e. the existence of a distant call of an unintended species) between the species' vocalisations is present, particularly in the mandrill recordings. Recordings can also contain noise from dogs, humans talking, or other human activities. The chimpanzees were recorded in two separate enclosures with two recorders per enclosure recording simultaneously. Hence, there may be overlap in vocalisations for recordings 1 and 2 as well as for recordings 3 and 4. This issue is considered in the chronological ordering based segmentation of the data into the training, validation and test sets. The total dataset amounts to a duration of 1112 hours, 358 GBs of original audio collected over a time span of 32 days.

\subsection{Annotation}
\label{subsec:anno}
\begin{figure}[htbp]
	\centering
	\includegraphics[width=\columnwidth]{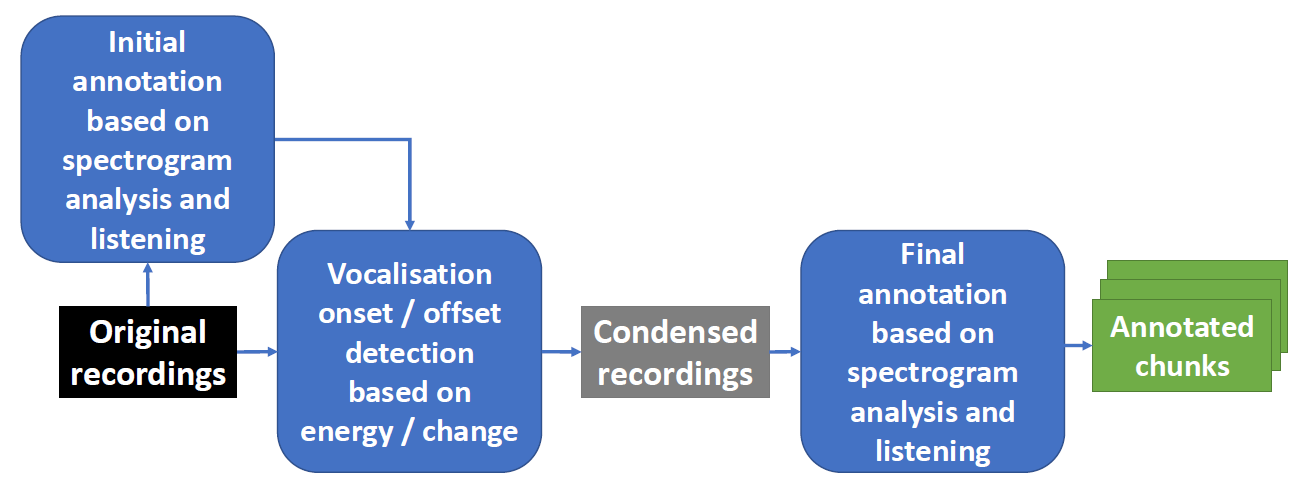}
	\caption{The semi-automatic annotation pipeline used in the study.}
	\label{fig:pp_anno_pipe}
\end{figure}

The first collection of annotations was compiled by manually reviewing the sound recordings and corresponding spectrograms in Raven Pro$\textsuperscript{\textregistered}$ software. To speed up this process, we `condensated' the data with an energy/change based automatic vocalisation detection using the first batch of manual annotations to estimate the detection performance. 
An overview of the semi-automatic annotation process is illustrated in Figure~\ref{fig:pp_anno_pipe}.
The detection comprises obtaining the power distribution from the power spectrum. From a species-specific frequency sub-band, we collect chunks (time-intervals) in which the registered signal loudness exceeds a species-specific threshold, or in which the local cumulative power distribution deviates from a global counterpart. The species specific thresholds are optimized to include close to all ($>$95\%) initial annotations and to remove as much background sections as possible.
The 'condensed' collection represents a set of timestamps, where we expect to hear disruptions in the ambient noise. The time-intervals are used to extract the corresponding signal fragments from our raw data. These fragments are bundled into a new audio file containing a high density of vocalisations that can be annotated more efficiently. 

\begin{figure}[htbp]
 \centering
 \begin{tabular}{c}

    Chimpanzee `Grunts'      \\
	\includegraphics[width=.98\columnwidth]{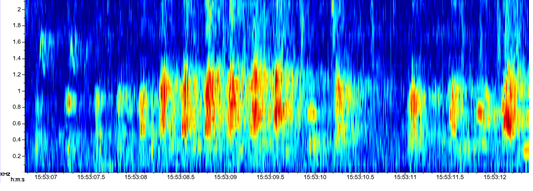}
      \\ 
      Chimpanzee `Pant Hoot' \\
     	\includegraphics[width=.98\columnwidth]{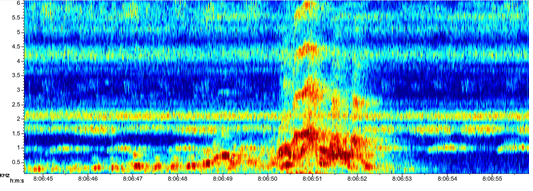}
      \\ 
      Chimpanzee `Screams' \\
     	\includegraphics[width=.98\columnwidth]{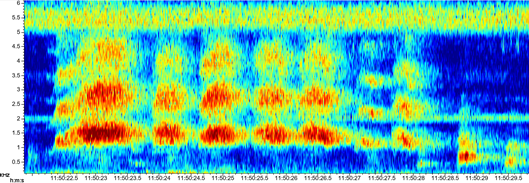}
      \\ 
      Mandrill `Two Phase Grunts' \\
     	\includegraphics[width=.98\columnwidth]{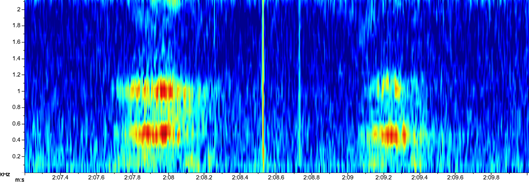}
      \\ 
      Red-Capped Mangabey `Wahoo' \\
     	\includegraphics[width=.98\columnwidth]{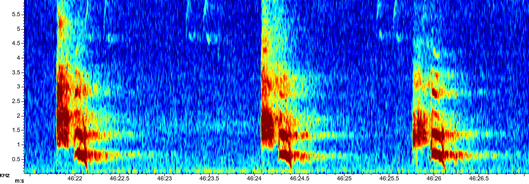}
      \\ 
      Red-Capped Mangabey `Whoop Gobble' \\
     	\includegraphics[width=.98\columnwidth]{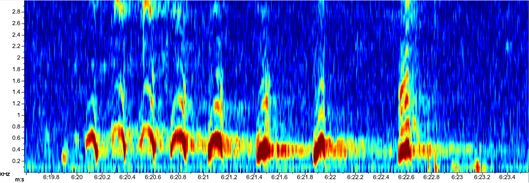}
      \\ 
        Guenon spp. `Pyow' \\
     	\includegraphics[width=.98\columnwidth]{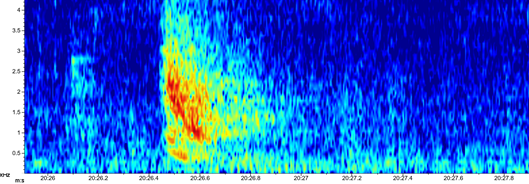}
      \\ 
 \end{tabular}
	\caption{Exemplar spectrograms for different vocalisations of the annotated primate species.}
    	\label{fig:call_types}
\end{figure}

Each species produces several vocalisation types, each varying in relative frequency, loudness and spectral properties. Experts consider these cues while observing the spectrogram (see Figure~\ref{fig:call_types} for exemplar spectrograms), spotting a candidate chunk and then listening to the selected chunk. This process yields over 10K annotated primate vocalisations with a class distribution of 6652 chimpanzee, 2623 mandrill, 627 red-capped mangabey and 476 of the mixed guenon group. 


\section{Benchmark Vocalisation Classification System}
To assess how well the species vocalisations can be automatically classified in the presented dataset, we present an acoustic primate classification system. The first stage is acoustic feature extraction, where we extract a standard set of acoustic descriptors from the signal and then summarize them using the statistical functionals (such as mean and standard deviation) over each chunk. This stage produces suprasegmental features of equal length. The next stage is machine learning, where the acoustic features and corresponding primate classes are input to a supervised learner. The details of these stages are given in the subsequent subsections.
\subsection{Acoustic Feature Extraction}
\label{subsec:acoustic_feats}
As acoustic Low-Level Descriptors (LLDs), we extract Mel-Frequency Cepstral Coefficients (MFCCs) 0-24 and Relative Spectral Transform (RASTA)~\cite{hermansky1994rasta} - Perceptual Linear Prediction (PLP)~\cite{hermansky1990perceptual} cepstrum for 12$^{th}$ order linear prediction, together with their first and second order temporal  coefficients ($\Delta$ and $\Delta\Delta$), making an LLD vector of 114 dimensions. The descriptors are then summarized using 10 functionals, based on the success observed in former paralinguistic studies~\cite{cciftcci2018turkish,KayaetalAVEC19}. The functionals used are: mean, standard deviation, slope and offset from the first order polynomial, the curvature (the leading coefficient) from the second order polynomial fit to the LLD contour, minimum value and its relative position, maximum value and its relative position, zero crossing rate of the LLD contour normalized into  [-1,1] range. This process yields $114 \times 10 = 1140$ supra-segmental acoustic features for each chunk, regardless of the number of frames. 

\subsection{Model Learning}
\label{subsec:classifiers}
In our work, we employ Kernel Extreme Learning Machine (ELM)~\cite{huang2012extreme} method, since this is a fast and accurate algorithm that previously produced state-of-the-art results on several paralinguistic problems~\cite{kaya2016fusing, kaya2017introducing}. 

Here, we opt to provide a brief explanation of ELM. Initially, ELM is proposed as a fast learning method for Single Hidden Layer Feedforward Networks (SLFN): an alternative to back-propagation~\cite{huang2004extreme}. To increase the robustness and the generalisation capability of ELM, a regularisation coefficient $\mathbf{C}$ is included in the optimisation procedure. Therefore, given a kernel $\mathbf{K}$ and the label vector $\mathbf{T} \in \mathbb{R}^{N \times 1}$ where $N$ denotes the number of instances, the projection vector $\mathbf{\beta}$ is learned as follows~\cite{huang2012extreme}:
\begin{equation}
\mathbf{\beta}  = (\frac{\mathbf{I}}{C}+\mathbf{K})^{-1} \mathbf{T}.
\end{equation}

In order to prevent parameter over-fitting, we use the linear kernel $\mathbf{K}(x,y) = x^Ty$, where $x$ and $y$ are the (normalised) feature vectors. With this approach, the only parameter of our model is the regularisation coefficient $C$, which we optimize on the validation set.

\section{Preliminary Experiments on the Primate vocalisation Dataset}
In this section we present our spectral analysis and the results of the preliminary classification experiments using the proposed benchmark system.

\subsection{Spectral Analysis of vocalisations}
\begin{figure}[htbp]
	\centering
	\includegraphics[width=.9\columnwidth]{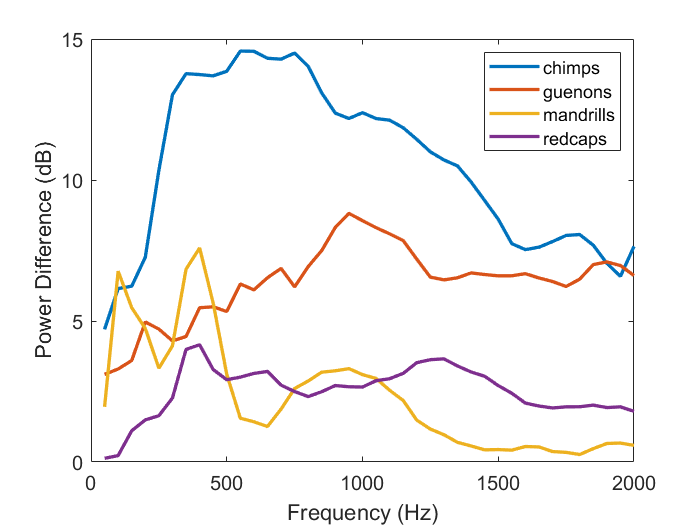}
	\caption{Average power (dB) difference between the mean vocalisation (signal) and background (noise) spectrum.}
	\label{fig:snr_primates}
\end{figure}

During the semi-automatic annotation process, we have analysed the spectral characteristics of vocalisations and the background noise per species. Based on domain knowledge and initial experimentation, we focused on spectral bands up to 2KHz. For this analysis, we have combined all annotated chunks for each primate class, obtained the power spectrum and then summarized the power in decibels (dB) using mean over time. We applied the same procedure for corresponding background portions for each species. The difference between the two means (see Figure~\ref{fig:snr_primates}) in dB provides an idea about the signal-to-noise ratio (SNR) and as such the relative difficulty of distinguishing each species' vocalisations in the given acoustic background conditions. In the figure, we observe multiple modes for mandrills and red-capped mangabeys, which correspond to different call types (c.f. Figure~\ref{fig:call_types}). In line with the acoustic observations during the annotations, vocalisations from mandrills and red-capped mangabeys have lower SNR values, making both the annotation and automated detection a harder problem.

\subsection{Classification Results}
For the classification experiments, we partitioned the dataset into training, validation and test sets using a temporal ordering (i.e. training correspond to the oldest, test to the newest recordings) with a ratio of 3:1:1, respectively. We set up two classification tasks 1) four-class classification of the species, 2) the four species classes plus the background chunks from the recordings of all species as the fifth class. To generate the background chunks, we sampled from the recordings not annotated as vocalisation, to exactly match the duration distribution of the annotated chunks of each species. This makes the five class problem highly imbalanced, as half of the chunks are of background class. However, such an imbalance is not extra-ordinary, if the final aim is to train classifiers for wildlife monitoring.

The models are trained on the training set, optimizing the Kernel ELM complexity hyper-parameter on the validation set. Then using the optimal hyper-parameter, the combination of the training and the validation sets are re-trained, and the corresponding model's predictions are checked against the ground truth test set labels. We use both accuracy and unweighted average recall (UAR), to report the predictive performance.

Using the acoustic features described in Section~\ref{subsec:acoustic_feats}, we then trained the Kernel ELM models with z-normalisation (ZN - standardising each feature such that they have zero mean and unit variance) and a combination of ZN with feature-vector level L2 normalisation, as suggested in~\cite{kaya2016robust}. When used with a linear kernel, L2  normalisation effectively converts the linear kernel into a cosine similarity kernel. The hyper-parameters of Kernel ELM method is optimized in the set $10^{\{-6,-5,-4,-3,-2,-1,0,1\}}$ with ZN and in in the set $10^{\{-1,0, 1, 2, 3, 4, 5, 6\}}$ with ZN+L2 normalisation combination. The respective validation and test performance of the trained models are summarized in Table~\ref{tab:perf_summary}. Note that we optimize for UAR due to class imbalance, while reporting both accuracy and UAR measures.
\begin{table}[htbp]
  \centering
  \caption{Validation and test set performances of KELM models for four and five-class classification tasks.}
    \begin{tabular}{llcccc}
    \toprule
          &       & \multicolumn{2}{c}{Validation} & \multicolumn{2}{c}{Test} \\
    \midrule
    Task  & Norm  & \multicolumn{1}{l}{Accuracy} & \multicolumn{1}{l}{UAR} & \multicolumn{1}{l}{Accuracy} & \multicolumn{1}{l}{UAR} \\
    \midrule
    \multirow{2}[1]{*}{Four-cls} & ZN    & 0.554 & 0.697 & 0.735 & 0.821 \\
          & ZN+L2 & 0.595 & 0.705 & 0.767 & 0.823 \\ \hline
    \multirow{2}[1]{*}{Five-cls} & ZN    & 0.603 & 0.610 & 0.682 & 0.707 \\
          & ZN+L2 & 0.617 & 0.627 & 0.697 & 0.698 \\
    \bottomrule
    \end{tabular}%
  \label{tab:perf_summary}%
\end{table}%

From the table, we observe that the test set (single probe for each normalisation option and task combination) performances are always better than the corresponding validation set performance. Moreover, all results are dramatically higher than chance-level UAR, which is 0.25 for the four-class and 0.2 for the five-class classification task. The results show that 1) the collected acoustic recordings have clear distinction for automatic discrimination of primate vocalisations, and 2) the proposed system has a good generalisation, reaching test set UAR scores of 0.82 and 0.70 in four-class and five-class classification tasks, respectively.

\section{Discussion and Conclusions}
\label{sec:conc}
 Initial results showed that we attain relatively high classification performance using our proposed system combining functionals of MFCC and RASTA-PLPC descriptors and modeling them using Kernel ELM. Data condensation also proved to be a valuable addition to the workflow for reducing the annotation workload. Our future aim is to apply the model on acoustic recordings of natural forests.
 
  Natural forest sounds pose the additional challenge of containing far fewer vocalisations compared to the sanctuary, and significantly higher levels of background noise, in particular in less relevant frequency bands. Moreover, similar to humans, primates can have varying vocal behavior across sex and age, including sex-specific call types, differences in frequency of specific vocalisation types, and differences in acoustic structures of shared call types~\cite{soltis2005african,mielke2013method}. There is also some extent of inter-individual variation, especially for chimpanzees ~\cite{fedurek2016sequential}. Considering the limited group sizes from which we derive our data, such variation may inevitably result in low generalisation when applied to the natural variation of individuals and group composition. Finally, not all species and species call types will be equally suitable for automated detection. Louder species such as chimpanzees will be more easily distinguished from background noise than for instance mandrills, and will consequently also have wider detection areas. Chimpanzees, however, often scream simultaneously, making it difficult to distinguish separate calls. 
  
  Future work lies in overcoming these challenges, which are partly caused because of the mismatch of acoustic conditions between sanctuary and natural data. Nonetheless, using sanctuary data has the advantage to provide relatively low-cost and accessible training data for classifiers, which may in turn boost the development and increased adoption of semi-automatic acoustic wildlife monitoring methods. To aid this development, the presented dataset is made publicly available in the context of the Interspeech 2021 Computational Paralinguistics Challenge (ComParE 2021)~\cite{compare2021}.
 
 \section{Acknowledgements}
\label{sec:ackn}
This research was funded by the focus area Applied Data Science at Utrecht University, The Netherlands.

\bibliographystyle{IEEEtran}

\bibliography{IS2020Primates}


\end{document}